\documentclass[11pt]{article}
\pdfoutput=1

\PassOptionsToPackage{hyphens}{url}
\usepackage[american]{babel}
\usepackage[bottom]{footmisc}
\usepackage[pdftex]{graphicx}
\usepackage[final]{acl}
\usepackage{times}
\usepackage{type1cm}
\usepackage[T1]{fontenc}
\usepackage[utf8]{inputenc}
\usepackage{microtype}

\usepackage{amsmath}
\usepackage{tabularx}
\usepackage{multirow}
\def\mmiddle#1{\mathrel{}\middle#1\mathrel{}}

\usepackage{url}
\RequirePackage{color}
\RequirePackage{soul}
\setstcolor{blue}

\usepackage{booktabs}

\DeclareGraphicsExtensions{.pdf,.ai,.jpg,.png}
\setkeys{Gin}{pagebox=artbox}
\graphicspath{{../arr21-topic-ontologies-for-argumentation-figures/}}

\usepackage{etoolbox}
\makeatletter
\patchcmd\@combinedblfloats{\box\@outputbox}{\unvbox\@outputbox}{}{%
}%
\makeatother

\begin{document}

\title{Modeling Proficiency with Implicit User Representations}

\newcommand{\darmstadt}{\textsuperscript{$\dagger$}}
\newcommand{\marburg}{\textsuperscript{$\ddagger$}}
\newcommand{\leipzig}{\textsuperscript{§}}

\author{%
Kim Breitwieser \darmstadt
\qquad Allison Lahnala \darmstadt
\qquad Charles Welch \darmstadt \\[1.5ex]
\bfseries Lucie Flek \marburg \hspace{1.5ex}
\bfseries Martin Potthast \leipzig \\[1.5ex]
\hspace{-5pt}\darmstadt{}TU Darmstadt,\quad\marburg{}University of Marburg,\quad\leipzig{}Leipzig University}

\date{}

\maketitle

\begin{abstract}
We introduce the problem of proficiency modeling: Given a user's posts on a social media platform, the task is to identify the subset of posts or topics for which the user has some level of proficiency. This enables the filtering and ranking of social media posts on a given topic as per user proficiency. Unlike experts on a given topic, proficient users may not have received formal training and possess years of practical experience, but may be autodidacts, hobbyists, and people with sustained interest, enabling them to make genuine and original contributions to discourse. While predicting whether a user is an expert on a given topic imposes strong constraints on who is a true positive, proficiency modeling implies a graded scoring, relaxing these constraints. Put another way, many active social media users can be assumed to possess, or eventually acquire, some level of proficiency on topics relevant to their community. We tackle proficiency modeling in an unsupervised manner by utilizing user embeddings to model engagement with a given topic, as indicated by a user's preference for authoring related content. We investigate five alternative approaches to model proficiency, ranging from basic ones to an advanced, tailored user modeling approach, applied within two real-world benchmarks for evaluation.
\end{abstract}

\section{Introduction}

\begin{figure}[!t]
\centering
\includegraphics[width=\linewidth]{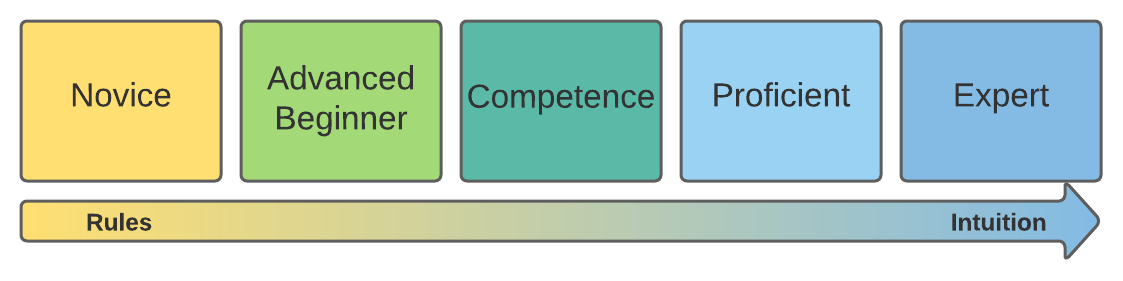}
\caption{Dreyfus model of skill acquisition, visualizing the process from rule-based to intuitive understanding of skill-related situations. Proficiency serves as a prerequisite for Expertise, which in turn represents mastery in a given skill.}
\label{fig:background:dreyfus}
\end{figure}

Social media is well-integrated into everyday life, providing a constant stream of content for users to consume. A user's stream is primarily composed of messages from people they follow, and, as that number increases, so does the number of posts presented to the user, until their capacity to consume each and every post is exceeded. Some platforms implement filtering criteria to omit posts that appear to be less relevant to the user (e.g., based on a combination of previous user engagement and the popularity of a post). Others simply display the most recent posts from the people the user follows at the time they start browsing. To the best of our knowledge, the skill level (see Figure~\ref{fig:background:dreyfus}) of a given followee is not taken into account. This paper introduces the problem of proficiency modeling, which is at the heart of solving such and similar problems of filtering and ranking by proficiency. Our contributions are threefold:
\begin{enumerate}
\item
Delineation of the problem of proficiency modeling based on existing psychological literature.
\item
Development of five proficiency modeling approaches, ranging from basic baselines to advanced models utilizing user embeddings.
\item
Evaluation of the five models within two real-world social media benchmarks involving celebrity posts on Twitter, and online mental health aid forums on Reddit.
\end{enumerate}

\section{Related Work}

Expertise is the last stage of skill development according to \citet{Dreyfus1980AFM}, which ranges from rote rule learning of novices to targeted training and practice in order to achieve expert level mastery as well as an intuitive understanding of how to solve new problems. However, to our knowledge, the modeling of skill levels {\em lower than} expertise is so far unexplored, whereas expert identification and retrieval have been studied even before the digital revolution. Since then, employers across all sectors have been interested in recognizing and measuring expertise, as well as deciding which candidates to pick given a specific task.

This need has only grown after the advent of the internet, providing us with information of orders of magnitude that are impossible to review manually, therefore requiring the aid of automated approaches to filter out relevant content \cite{balog2012expertise}. In general, these problems task a system with finding the most fitting users given a predefined topic, also known as expert finding \cite{van2016unsupervised}, as well as creating a structured overview containing fields of expertise given a user, also known as expert profiling \cite{10.5555/1625275.1625703}. In downstream applications, the obtained knowledge can be used in a variety of ways, such as expert recommendation, expert ranking, and expert clustering \cite{gonccalves2019automated}.

Early approaches employ basic heuristics, assuming that the last user to work on a given task can be considered a likely expert due to what's known as the recency effect, or by concentrating on a specific subset of documents \cite{mcdonald2000expertise,baddeley1993recency}. Newer traditional models handle heterogeneous datasets, making use of term frequencies \cite{DBLP:conf/sigir/BalogR07} and topic modeling techniques, such as LDA \cite{10.5555/944919.944937}, often in combination with additional features like social relationships between users \cite{xie2016topic}. More recently, neural networks have also been employed in order to retrieve and rank experts, often in the context of question answering \cite{zhao2016expert}. We adopt the basic models as baselines, and develop for the first time a user modeling approach for proficiency that gets by unsupervised.

\section{On the Importance of Proficiency}

Proficiency is defined as being \textit{"well advanced in an art, occupation, or branch of knowledge"}, with the act of becoming proficient being defined as the \textit{"advancement in knowledge or skill"}. Said knowledge is typically acquired through training and practical experience, resulting in a deep familiarity with the subject, though the exact degree of knowledge necessary to qualify as being proficient can seem arbitrary. Domain-specific knowledge measurable in terms of proficiency or expertise has been found to be largely unrelated to general intelligence or cognitive abilities, and rarely translates outside of its domain \cite{simon1973skill}.

Proficiency and expertise, while sometimes used interchangeably, are related terms with the latter signifying distinct mastery in a given area. On the Dreyfus model of skill acquisition (see Figure~\ref{fig:background:dreyfus}), proficiency and expertise represent individual stages, signifying the last steps in the process of acquiring, and finally mastering, skills \cite{Dreyfus1980AFM,dreyfus2000mind}. When comparing these related concepts to each other, it should be noted that expertise requires deliberate practice, a systematic approach focused on improving performance related to specific task-related aspects, while a generally high level of proficiency can be reached through a less focused training approach, aimed at the pure accumulation of task-related knowledge \cite{expertise_practice}. Proficiency, but not expertise, can therefore be approximated by a sustained, and active exposure of a user to a given topic, which can be frequently observed in all kinds of online forums.

Recognizing proficient people is especially important in group-related activities and communities, since, in the absence of experts, they are the next best reliable source of knowledge when making decisions, and their arguments have a higher weight (i.e., persuasiveness) in discussions \cite{group_decision_expertise,YUE20122900}. Being unsuccessful in recognizing proficiency usually results in process loss, often caused by the group not expending the necessary efforts to recognize proficient members, or being persuaded by less experienced, but dominant personalities. These individuals then tend to proceed to prioritize their own goals, which are often suboptimal when applied to the group as a whole \cite{aronson2014sozialpsychologie}.

\section{Modeling Proficiency}

Given that proficiency in a given topic results from an increased exposure to it, we work under the (heuristic) assumption that a given user's tendency to author topic-related content correlates with their personal exposure to the topic. Specifically, we define a set of $n$~query words $Q = \{w_1, ..., w_n\}$ related to the topic and use a variety of models to transform it into a set of features $F_u = \{f_{1, u}, ..., f_{n, u}\}$, which separately exists for each user~$u \in U$, where $U$ is the set of all users in a given community. These features are used in order to classify users with known areas of proficiency.

\subsection{User Representations}

The following five models are considered:

\paragraph{Term Frequency (TF)}
Given the set of query words, we measure the term frequencies in a user's post history, adjusted for the total number of words for that user to account for potentially unevenly distributed text quantities:
\begin{equation}
\text{TF}_u = \left\{ \frac{f_{w_i, u}}{n_u} \mmiddle| w_i \in Q \right\},
\end{equation}
where~$f_{w_i,u}$ denotes the number of appearances of word~$w_i$ in all posts authored by user~$u$, and~$n_u$ describes the total number of words across the user's entire post history, which remains static for all~$w_i \in Q$.

\paragraph{Term Frequency – Inverse Document Frequency (TF-IDF)}
The TF model only relies on the user being modeled, without taking into account the greater population being present in the dataset. TF-IDF aims to resolve this issue by additionally measuring the word frequencies over the whole dataset \cite{Jones72astatistical}. We apply this technique by counting the number of user post histories containing the given word and setting them into relation to the total number of users present in the dataset:
\begin{equation}
\text{IDF}_{w_i} = \log \frac{|U|}{1 + \left|\{ f_{w_i, u} \neq 0 \mmiddle| u \in U \right\}|},
\end{equation}
where~$|U|$ denotes the total number of users in the dataset, and $\left|\{ f_{w_i, u} \neq 0 | u \in U \right\}|$ the subset of users having authored at least one post containing~$w_i$, offset to account for possible zero division errors. The final model $\text{TF-IDF}_u$ is obtained by multiplying the respective vectors $\text{TF}_u$ and $\text{IDF}_u$ of corresponding scores per~$w_i\in Q$.

\paragraph{User2Vec (U2V)}
We represent users by their User2Vec embeddings \cite{amir2016modelling}, which are themselves a form of paragraph vectors \cite{DBLP:journals/corr/LeM14}, treating all content authored by a given user as a single document. We obtain these embeddings using pretrained $400$-dimensional Word2Vec embeddings trained on data originating from Twitter \cite{godin2019}. User2Vec embeddings are trained with the objective of maximizing the probability of a user's sentences to be assigned to that user. In order to obtain these probabilities, word and user embeddings are combined with each other in a non-linear way:
\begin{equation}
\text{U2V}_u = \left\{ \sigma \left( \mathbf{w_i} \cdot \mathbf{u_u} \right) \mmiddle| w_i \in Q \right\},
\end{equation}
where~$\sigma$ denotes the Sigmoid activation function, $\mathbf{w}_i$~the Word2Vec representation of the query term~$w_i$, and~$\mathbf{u}_u$ is the User2Vec representation of the given user~$u$, trained using $15$~negative samples per word with an initial learning rate of~$0.00005$, exponentially decaying by a factor of~$10^{-1}$ whenever the model improves. We perform this training process over a maximum of $25$~epochs, potentially stopping early after $5$~epochs without improvement (patience).

Unlike the previously introduced models, this model also extends to semantically related words due to the usage of word embeddings compared to the words themselves. This includes word variations, such as plurals, which should allow the model to make use of a wider variety of features compared to preexisting baselines.

\paragraph{Relative User2Vec (Rel-U2V)}
Similarly to TF-IDF, Rel-U2V extends the U2V model by taking into account all users in the dataset, obtaining the average U2V scores across all users and setting them into relation to the user currently being evaluated:
\begin{equation}
\text{Rel-U2V}_u = \left\{ \frac{\text{U2V}_{u, w_i} \cdot |U|}{\sum_{u_k \in U} \text{U2V}_{u_k, w_i}} \mmiddle| w_i \in Q \right\}.
\end{equation}

The values obtained this way are centered around~$1$, with higher values indicating an above-average probability for the given user to be proficient on $Q$'s topic, and vice versa, while penalizing words with a high probability across the whole dataset, such as stop words and terms with a low amount of user-specific information.

\paragraph{Latent Dirichlet Allocation (LDA)}
While not being based on the query terms~$Q$ used to evaluate the other models, LDA \cite{10.5555/944919.944937} is an unsupervised topic modeling approach that could potentially be used to model proficiency as well. It models the probability $P(w, t)$ of a given word $w$ to belong to topic $t$ as follows:
\begin{equation}
P(w, t) = P(t|d) \cdot P(w|t), 
\end{equation}
where $d$ is a document in the entire dataset, or in our case a post authored by a user. By averaging over the posts and words authored by a given user $u$, these scores can then be transformed from a word to a user level, resulting in a set of features $F_u = \left\{ f_{t, u} \mmiddle| t \in M \right\}$ with $M$ representing the LDA~model itself.

\subsection{Scoring Proficiency}

After obtaining $F_u$ using one of the models outlined in the previous section, we use these to fit a Support Vector Machine~(SVM) \cite{cortes1995support}, using the known proficiency topics for each user as a ground truth. In doing so, we are able to classify users based on their proficiencies.

In addition, by averaging the features obtained using query-based models, we obtain a scalar value, which we refer to as Proficiency Score~(PS):
\begin{equation}
\text{PS}_{u, c} = \frac{\sum_{f_{i,u} \in F_u} f_{i,u}}{|F_u|}.
\end{equation}

We can apply this score to any combination of user~$u$ and content~$c$, using it as a measure of how proficient the user in question is regarding the proficiency topics contained in a piece of content. We explore the usefulness of these scores based on the proficiency models U2V and Rel-U2V.

\section{Datasets}

We evaluate our approach on two datasets, both annotated in terms of user occupations, which we interpret to be topics of proficiency for them (among others). Beforehand, the datasets are preprocessed using the following pipeline:
\begin{enumerate}
\item
Whitespace normalization (e.g., collapsing a sequence of spaces to a single space), allowing for easier tokenization.
\item
Conversation to lowercase. Especially on social media, word casing often varies between posts and users, otherwise complicating the lookup of word embeddings.
\item
Repeated character reduction. In English, repeated character text is kept at a maximum of~2, with 3~letters only occurring on compound words. Even so, social media content can contain artificially elongated words, which we reduce to a common form to ease the lookup.
\item
Masking of user mentions with \textit{@user}. Content on social media platforms like Twitter or Reddit can refer to other users using the \textit{@}-prefix. User names are usually not part of pretrained embeddings, which is why we replace all occurrences with a standardized placeholder.
\item
Masking of hyperlinks and (phone) numbers with placeholders, for the same reasons.
\end{enumerate}

\subsection{\textsc{MHP} dataset}

This dataset, originally collected from so-called ``Subreddits'' of the social media platform Reddit, consists of posts and comments authored in environments discussing mental health issues \cite{DBLP:journals/corr/abs-2106-12976}. In order to ensure enough content to properly model each user, we further filter the dataset to only include users with at least 100~posts and/or comments, leaving us with a total of 5,299 users.

\begin{table}[t]
\small
\centering
\begin{tabularx}{\linewidth}{@{}X@{}}
\toprule
\bfseries Query $Q$ \\
\midrule
abuse, acceptance, addiction, adhd, advice, affective, al, alcohol, alcoholism, anger, anxiety, ask, aspergers, bipolar, body, bpd, bulimia, compulsive, cope, coping, counseling, crippling, dbt, depression, disability, disorder, diversity, docs, domestic, drinking, dysmorphic, eating, emetophobia, fap, friend, gfd, harm, health, help, illness, kratom, leaves, mental, neuro, neurodiversity, ocd, pcos, picking, psychiatry, ptsd, rape, recovery, relationship, sad, schizophrenia, self, skills, skin, smoking, social, stop, suicide, therapist, therapy, tourette, trauma, violence \\
\bottomrule
\end{tabularx}
\caption{Query words used for the differentiation of MHPs and peer users in the \textsc{MHP} dataset. The list was obtained from the names of Subreddits used to construct the dataset.}
\label{tab:datasets:mhp:query}
\end{table}

In the context of Reddit, posts originally start a conversation, while comments are authored in response to them. To create our embeddings, we do not differentiate between the different content types.

The dataset is split between mental health professionals (MHPs) and peers, namely users without proficiency in the medical domain, asking for advice. Since the dataset is unbalanced with only 262~(4.94\%) users being labeled as~MHPs, our evaluation is performed twice, once on the full dataset and again on a balanced subset containing all MHPs and an equal number of randomly selected peers.

Apart from the user labels themselves, the dataset also provides a list of Subreddits used to collect the data. We use these names, filtered for those having a matching representation in the pretrained Word2Vec embeddings, in order to construct the query~$Q$ being used to construct the features to evaluate our models on. The complete list of resulting terms can be found in Table~\ref{tab:datasets:mhp:query}.

\subsection{\textsc{Celebrities} dataset}

This dataset, originally collected from the social media platform Twitter, consists of verified human user accounts deemed notable enough to receive their own Wikipedia page and associated Wikidata entry \cite{wiegmann-etal-2019-celebrity}. This entry was used to collect metadata about the users, including a list of occupations, which we used as a stand-in for the user's areas of proficiency. Similarly to~\textsc{MHP}, this dataset was also filtered, this time for users with at least 1,000~posts, due to the smaller size of Twitter posts as compared to content posted on Reddit. Out of these users, we furthermore selected all occupations with at least 500~users and grouped those together to 5~areas of proficiency, described in detail in Table~\ref{tab:datasets:celebrities:classes}. Finally, we restricted the dataset to users assigned to at least one of these areas, leaving us with a total of 28,566~users. 

\begin{table}[t]
\small
\renewcommand{\arraystretch}{1.06}
\centering
\begin{tabularx}{\linewidth}{@{}l r X@{}}
\toprule
\bfseries Proficiency &\bfseries Users &{\bfseries Occupations} (Wikidata ID) \\
\midrule
politics & 4161 & politician (Q82955) \\
\midrule
sports & 8219 & basketball player (Q3665646), baseball player (Q10871364), association football player (Q937857), American football player (Q19204627), ice hockey player (Q11774891), rugby union player (Q14089670) \\
\midrule
journalism & 3460 & journalist (Q1930187) \\
\midrule
music & 6549 & singer (Q177220), composer (Q36834), guitarist (Q855091), songwriter (Q753110), record producer (Q183945), singer-songwriter (Q488205), musician (Q639669) \\
\midrule
acting & 8329 & actor (Q33999), film actor (Q10800557), television actor (Q10798782), voice actor (Q2405480), stage actor (Q2259451) \\
\bottomrule
\end{tabularx}
\caption{Assigning user occupations to proficiencies for the \textsc{Celebrities} dataset. As users can be assigned to more than one proficiency topic, the added user counts end up being higher than the number of users in the dataset.}
\label{tab:datasets:celebrities:classes}
\end{table}

For each of these areas, we choose a number of words according to their frequencies in the post histories of related users to build our query~$Q$, which can be seen in Table~\ref{tab:datasets:celebrities:query}. It should be noted that the same query was used across all tasks, enabling both binary classification for a given area of proficiency, as well as multilabel classification across all of them, which is especially useful considering a portion of the users are labeled with more than a single proficiency.

\begin{table}[t]
\small
\renewcommand{\arraystretch}{1.06}
\centering
\begin{tabularx}{\linewidth}{@{}l X@{}}
\toprule
\bfseries Proficiency & \bfseries Query $Q$ \\
\midrule
politics & administrative, constituency, directive, election, governance, government, opposition, ratification, referendum, reform \\
\midrule
sports & \#championsleague, \#europaleague, \#premierleague, baseball, basketball, college, fifa, football, foul, game, goal, hockey, mvp, nba, nfl, nhl, playoff, rugby, score, touchdown, walkoff, win \\
\midrule
journalism & allegation, cite, debunk, informant, left, refute, right, scandal, tabloid, unfounded, whistleblower \\
\midrule
music & acoustic, album, cd, discography, instrumental, mixtape, music, remix, riff, self-titled, singer, song, tracklist, vocals \\
\midrule
acting & actor, actress, cast, cosplay, co-star, finale, miniseries, movie, premiere, prequel, sequel, spinoff \\
\bottomrule
\end{tabularx}
\caption{Query words used for the differentiation of different proficiency topics in the \textsc{Celebrities} dataset. For better readability, words are grouped by their respective class, though for the evaluation only the combined set of all words was used. The list was obtained based on word frequencies for users of listed proficiency topics.}
\label{tab:datasets:celebrities:query}
\end{table}

\section{Experimental Results}

In order to evaluate the models, we compute the features~$F_u$ for each user in the dataset and the words present in a query~$Q$. We then use these features to fit a Support Vector Machine (SVM) \cite{cortes1995support} and measure its accuracy and F1~score, both for binary and---in case of the \textsc{Celebrities} dataset---multilabel classification, with the ground truth being set to the proficiency topics defined over one or multiple occupations, as outlined in the previous section. When performing multilabel classification, we report the micro-averaged F1~score over individual users. Each evaluation is performed and averaged over $5$~folds, stratified on the previously described proficiency topics. 

\begin{table}[t]
\small
\centering
\begin{tabularx}{\linewidth}{@{}l X c c@{}}
\toprule
\bfseries Dataset & \bfseries Model & \multicolumn{1}{@{}c@{}}{\bfseries Acc (\%)} & \multicolumn{1}{@{}c@{}}{\bfseries F1} \\
\midrule
\multirow{4}{*}{full}     & \textbf{TF}      & 96.89 $\mp$ 0.26 & 0.59 $\mp$ 0.06 \\
                          & \textbf{TF-IDF}  & 96.51 $\mp$ 0.31 & 0.53 $\mp$ 0.06 \\
                          & \textbf{U2V}     & 96.62 $\mp$ 0.25 & 0.51 $\mp$ 0.07 \\
                          & \textbf{Rel-U2V} & 96.60 $\mp$ 0.25 & 0.51 $\mp$ 0.06 \\
\midrule
\multirow{4}{*}{balanced} & \textbf{TF}      & 86.08 $\mp$ 3.72 & 0.85 $\mp$ 0.04 \\
                          & \textbf{TF-IDF}  & 86.65 $\mp$ 3.22 & 0.86 $\mp$ 0.03 \\
                          & \textbf{U2V}     & 85.11 $\mp$ 2.31 & 0.85 $\mp$ 0.03 \\
                          & \textbf{Rel-U2V} & 85.68 $\mp$ 2.50 & 0.85 $\mp$ 0.03 \\
\bottomrule
\end{tabularx}
\caption{Accuracy and F1~scores obtained on the \textsc{MHP} dataset when fitting the same SVM architecture with different model features obtained on the same set of query terms. Tests were performed both on the full dataset, as well as on a balanced subset containing an equal amount of users labeled as MHPs and peers. All results were obtained using 5-fold cross-validation and report the mean value as well as standard deviation.}
\label{tab:evaluation:mhp}
\end{table}

\subsection{\textsc{MHP} Dataset Results}

The results for the \textsc{MHP} dataset can be found in Table~\ref{tab:evaluation:mhp}. As can be seen in both the Accuracy and F1~score columns, all obtained results are comparable with each other, considering the standard deviations. This is especially true on the balanced subset of data, likely due to the resulting small size of the dataset, thus providing not enough information to properly fit the SVM. This can be attributed to the fact that the used query terms are by themselves quite exhaustive, so the added coverage due to the usage of embeddings is not necessary in order to obtain the same level of result quality as when directly using the words themselves. Though it can be observed that all evaluated models are generally lacking in terms of recall, which we attribute to the fact that the used query terms are mainly aimed at identifying MHPs, and not peers, as well as the Subreddits used to obtain the dataset itself generally being aimed at mental health, increasing the likelihood of non-proficient users to make use of domain-related vocabulary.

\begin{table}[t]
\small
\centering
\begin{tabularx}{\linewidth}{@{}l X c c@{}}
\toprule
\bfseries Scenario & \bfseries Features & \multicolumn{1}{@{}c@{}}{\bfseries Acc (\%)} & \multicolumn{1}{@{}c@{}}{\bfseries F1} \\
\midrule
\multirow{4}{*}{multilabeling} & \textbf{TF}      & 63.25 $\mp$ 4.88 & 0.75 $\mp$ 0.02 \\
                               & \textbf{TF-IDF}  & 60.70 $\mp$ 4.85 & 0.73 $\mp$ 0.03 \\
                               & \textbf{U2V}     & 74.70 $\mp$ 2.35 & \textbf{0.82 $\mp$ 0.02} \\
                               & \textbf{Rel-U2V} & \textbf{74.75 $\mp$ 2.32} & \textbf{0.82 $\mp$ 0.02} \\
                               & \textbf{LDA}     & 49.69 $\mp$ 2.81 & 0.70 $\mp$ 0.01 \\
\cmidrule[2pt]{1-4}
\multirow{4}{*}{politics}      & \textbf{TF}      & 93.37 $\mp$ 0.72 & 0.74 $\mp$ 0.04 \\
                               & \textbf{TF-IDF}  & 92.21 $\mp$ 0.56 & 0.68 $\mp$ 0.03 \\
                               & \textbf{U2V}     & 96.35 $\mp$ 0.55 & \textbf{0.87 $\mp$ 0.02} \\
                               & \textbf{Rel-U2V} & \textbf{96.35 $\mp$ 0.52} & \textbf{0.87 $\mp$ 0.02} \\
                               & \textbf{LDA}     & 95.32 $\mp$ 0.23 & 0.76 $\mp$ 0.02 \\
\midrule
\multirow{4}{*}{sports}        & \textbf{TF}      & 92.24 $\mp$ 1.96 & 0.86 $\mp$ 0.03 \\
                               & \textbf{TF-IDF}  & 91.64 $\mp$ 2.00 & 0.85 $\mp$ 0.04 \\
                               & \textbf{U2V}     & 93.53 $\mp$ 1.04 & \textbf{0.88 $\mp$ 0.02} \\
                               & \textbf{Rel-U2V} & 93.51 $\mp$ 1.07 & \textbf{0.88 $\mp$ 0.02} \\
                               & \textbf{LDA}     & \textbf{94.03 $\mp$ 1.17} & 0.76 $\mp$ 0.05 \\
\midrule
\multirow{4}{*}{journalism}    & \textbf{TF}      & 87.90 $\mp$ 0.02 & 0.01 $\mp$ 0.01 \\
                               & \textbf{TF-IDF}  & 87.89 $\mp$ 0.00 & 0.00 $\mp$ 0.00 \\
                               & \textbf{U2V}     & 92.14 $\mp$ 0.60 & \textbf{0.62 $\mp$ 0.03} \\
                               & \textbf{Rel-U2V} & \textbf{92.17 $\mp$ 0.60} & \textbf{0.62 $\mp$ 0.03} \\
                               & \textbf{LDA}     & 90.23 $\mp$ 0.35 & 0.37 $\mp$ 0.06 \\
\midrule
\multirow{4}{*}{music}         & \textbf{TF}      & 91.24 $\mp$ 1.18 & 0.70 $\mp$ 0.06 \\
                               & \textbf{TF-IDF}  & 90.86 $\mp$ 1.13 & 0.77 $\mp$ 0.04 \\
                               & \textbf{U2V}     & \textbf{92.20 $\mp$ 0.35} & \textbf{0.81 $\mp$ 0.01} \\
                               & \textbf{Rel-U2V} & 92.20 $\mp$ 0.39 & \textbf{0.81 $\mp$ 0.01} \\
                               & \textbf{LDA}     & 80.71 $\mp$ 0.50 & 0.65 $\mp$ 0.01 \\
\midrule
\multirow{4}{*}{acting}        & \textbf{TF}      & 88.74 $\mp$ 1.67 & 0.78 $\mp$ 0.04 \\
                               & \textbf{TF-IDF}  & 87.68 $\mp$ 1.71 & 0.76 $\mp$ 0.04 \\
                               & \textbf{U2V}     & 90.87 $\mp$ 0.20 & \textbf{0.83 $\mp$ 0.01} \\
                               & \textbf{Rel-U2V} & \textbf{90.88 $\mp$ 0.18} & \textbf{0.83 $\mp$ 0.01} \\
                               & \textbf{LDA}     & 75.35 $\mp$ 2.33 & 0.76 $\mp$ 0.03 \\
\bottomrule
\end{tabularx}
\caption{Accuracy and F1~scores obtained on the \textsc{Celebrities} dataset when fitting the same SVM architecture with different model features obtained on the same set of query terms, except for LDA, whose features correspond to the 50 topics modeled by the underlying LDA~model. All results were obtained using 5-fold cross-validation and report the mean value as well as standard deviation.}
\label{tab:evaluation:celebrities}
\end{table}

\subsection{\textsc{Celebrities} Dataset Results}

The results for the \textsc{Celebrities} dataset can be found in Table~\ref{tab:evaluation:celebrities}. Tests were performed both in a multilabeling scenario accounting for users assigned to more than one field of proficiency, as well as binary classification for each proficiency individually. For multilabeling, F1 refers to the micro-averaged scores across individual users.

In addition to the previous models, we also include results obtained on an LDA~model, created from the 5,000~top users of the \textsc{Celebrities} dataset, based on the amount of posts authored, as a higher amount of data has proven to be computationally infeasible in our environment. We create the LDA~model by using a single pass over the data and a total of 50~LDA-topics (not to be confused with proficiency topics), which we empirically chose as a high number to not produce overlap between the LDA-topics. 

The results for the \textit{multilabeling} scenario vary strongly between models, with embedding-based implementations scoring noticeably higher than word-based ones. We attribute this to the increased complexity of the task, combined with the fact that the used query contains comparatively less entries per proficiency topic, profiting from the added coverage provided by the embeddings. These models also generally achieve better results on the binary classification tasks, which we also attribute due to the sparse query, but we want to highlight some of them in particular:

\begin{figure}[t]
\centering
\includegraphics[width=\linewidth]{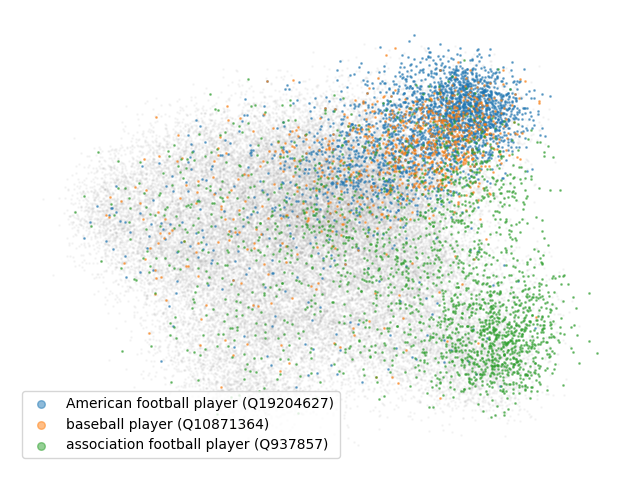}
\caption{2-dimensional PCA representation of User2Vec embeddings. Shown are users labeled as American football players, baseball players and association football (soccer) players, all of which share a common proficiency in the sports topic. The main clusters of these user groups are fairly close to each other, additionally separated between nationalities, as American football and baseball are traditionally more common in the USA.}
\label{fig:usr2vec:clusters}
\end{figure}

For the \textit{politics} scenario, the LDA~model achieves generally high results, which can be attributed to the presence of several LDA-topics dedicated to politics-related words. Though the results generally fall slightly short compared to embedding-based approaches, especially in terms of recall, which proves to be a general downside of the LDA~model based on the observed F1~scores.

For the \textit{sports} scenario, the LDA~model achieves the highest results in terms of accuracy, yet again lacks behind in recall, as expressed in a low F1~score compared to all other models being evaluated. Looking at the topics described by the LDA~model, we also could not find a specific LDA-topic correlating with sports, leading us the conclusion that, while an LDA-based approach can generally work to classify user proficiencies, it depends highly on the described LDA-topics and thus proves unreliable in order to model proficiency. We also observe that, in our case, the described models often correlate with syntactic classes, such as verbs describing motion or holidays, and rarely with areas of knowledge or experience.

For the \textit{journalism} scenario, the word-based approaches, including LDA, generally show a very low recall, likely due to the fact that journalistic content often refers to other topics, discerning itself by the usage of observing or describing sentence patterns. 

To further validate User2Vec's ability of grouping different yet closely related occupations of proficiency, Figure~\ref{fig:usr2vec:clusters} showcases users with different occupations, which are still part of the same proficiency topic, to be situated next to each other when being modeled this way. The clustering ability of user embeddings has been previously shown e.g. for the user's age, gender, and sarcasm tendency \cite{amir2016modelling,DBLP:journals/corr/AmirCCSW17,DBLP:journals/corr/abs-1712-03538}.

\begin{table}[t]
\small
\centering
\begin{tabularx}{\linewidth}{@{}l X r r@{}}
\toprule
\bfseries PS & \bfseries  Content \\
\midrule
1.78 & president obama is standing up for those who serve — join veterans and military families for obama to stand with him : url \\
\midrule
1.57 & president obama laid out a plan to restore the promise of middle class security last night : url \\
\midrule
1.66 & “ its going to empower students and families to be back in the drivers seat . ” — vp biden on new college financial aid tools \\
\midrule
1.09 & [ trump ] was born with a silver spoon in his mouth that hes now choking on because his foots in his mouth along with the spoon . - vp biden \\
\midrule
1.02 & the first corvette was \#madeinamerica ( flint , mi ) on this day in 1953 . url \\
\midrule
0.72 & happy easter ! url \\
\bottomrule
\end{tabularx}
\caption{Proficiency scores obtained on a variety of posts authored by US~president Joe Biden, highlighting the difference between content belonging to a given user's proficiency topics and unrelated content. The dataset in question was created in 2018, which is why proficiency-related content refers to past events.}
\label{tab:applications}
\end{table}

\subsection{Scoring Individual Posts by Proficiency}

So far we explored the possibility of using the computed scores in order to classify users based on proficiency topics, which themselves are represented by a set of query terms. We handpicked the terms based on word frequencies, but topic models such as LDA could also be applied in order to represent individual domains. The proposed model can also be used to measure the likelihood of a given post to be part of the author's areas of proficiency. Given the amount of information we are confronted with on social media, being able to gauge whether a given post is informative or consists of everyday chitchat can be useful in order to manage time and direct effort towards relevant content.

Proficiency can be scored in different ways, depending on the proficiency model used to compute~PS:
(1)~Proficiency related to the user itself, using the U2V model and treating individual users independently from each other. These scores specifically prioritize content that’s highly characteristic for the user and should not be set into comparison with other users, but can provide a useful tool in order to filter out content unrelated to the proficiencies of a given user.
(2)~Using the Rel-U2V model, modeling proficiency in relation to the overall population. This approach can be utilized as a form of comparison, for example to threshold users based on their proficiency scores related to a given topic.

Table~\ref{tab:applications} contains examples of this application, showcasing the scoring of different posts authored by the current US~president Joe Biden. In this specific example, the proficiency score~PS was computed using the Rel-U2V model. A fully-fledged evaluation of proficiency scoring of individual posts of individual users, however, requires ground truth labels that reveal which posts of which user are within their domains of proficiency or without. A third-party assessment of judging whether a user is proficient in a given topic and whether a post reflects this proficiency, requires judges who intimately know the people in question---which, for celebrities, is not beyond reason, but still beyond the scope of our research. In future work, we plan to compile a corresponding benchmark to enable the study of proficiency scoring at the post-level.

\section{\mbox{Design Variations of User2Vec\,Models}}

While the User2Vec-based models we propose and evaluate seem promising, especially in situations when the available query data is sparse and can not be directly matched, there are several possible modifications to these approaches.

\subsection{Capturing Positional Information}

As User2Vec embeddings are created using a bag-of-words approach, the grammatical structure of a sentence is lost. Therefore, the sentences \textit{"Dog bites man"} and \textit{"Man bites dog"} would be considered identical, even though they do not have the same meaning.
This might make the User2Vec proficiency approach vulnerable to buzzword dropping. Expanding the model to longer word n-gram spans leads to more robustness against this issue.

\subsection{Robustness to Varying Sentence Length}

When computing the proficiency score~PS, we average model features obtained using User2Vec embeddings, causing the resulting scores to favor short sentences only containing a few, but highly probable words. This problem is mitigated by using BLEU's brevity penalty and specifying a reference length~$r$, either statically or dynamically inferred. Shorter sentences will then be penalized, causing the scoring function to heavily reduce their score, effectively discarding those that are too far below the reference \cite{papineni-etal-2002-bleu}:
\begin{equation}
\hat{\text{PS}_{u, c}} = \exp \left( \min \left( 0, 1 - \frac{r}{|c|} \right) \right) \cdot \text{PS}_{u, c},
\end{equation}
with~$|c|$ being the length of the piece of content being scored. When computing scores on posts authored by a given user, the average length of this user's posts serves as a good reference point for~$r$, making sure posts need to both cover the user's area of proficiency, as well as providing a similar amount of information as is usual for this particular user.
However, the content length is not guaranteed to correlate with the amount of information present in all cases, and, given that the brevity penalty in BLEU is added to aid in the quality-evaluation of machine-generated text, while our datasets consist solely of human-authored posts, we found the addition to rarely be helpful. Therefore, we suggest to evaluate the usefulness of penalizing brevity in the context of given dataset characteristics.

\subsection{Anonymization in Preprocessing}

We initially preprocessed the datasets used during our evaluation by masking user mentions with a unified \textit{@user} placeholder. However, mentions can often contain valuable information, relating them to proficiency topics or providing context. As an example, the post
\begin{quote}
\textit{I can ’ t emphasize enough how @user would not be where it is today without @user Thank you}
\end{quote}
authored by Elon Musk (@elonmusk) on Twitter originally refers to the companies of \textit{SpaceX} and \textit{NASA}, respectively. By replacing these mentions with a unified placeholder, the domain information is lost. For the right balance between accuracy and robustness, we propose removing the leading~\textit{@}, which will allow the usage of the specific embedding, given that the mentioned entity is known widely enough. A hybrid approach is possible here - first performing a lookup to test for the presence of the mentioned entity and, in case one couldn't be found, defaulting to a unified placeholder.

\section{Ethical and Privacy Considerations}

Given that both the introduced problem definition, as well as the models proposed in order to tackle it, aim at the labeling of users, as well as extending to a form of rating based on computed proficiency, there are certain ethical implications involved with classifying users, both for correctly made statements as well as potential misclassifications \cite{rudman2012social}. We therefore strongly advise to not use the proposed models as the sole basis for decisions made concerning the fate of humans, such as to which candidates to pick given a certain position.

As for the topic of privacy, all data used by us in the creation of our models, as well as subsequent evaluation, are publicly available on the Twitter and Reddit social media platforms. It should be noted that the Developer Agreements of such platforms generally explicitly forbid the usage of data obtained through the platform for the purpose of surveillance or in order to perform discriminatory actions, as exemplary outlined in \citet{Pardo2013OverviewOT}. We therefore explicitly state, that the scope of this work is limited to the evaluation of models based on publicly available data in order to approach the problem of proficiency modeling, and not in case used to discriminate or surveil individual users based on the informations obtained.

\section{Conclusion}

We introduced the problem of proficiency modeling in the scope of social media, aimed at both identifying the proficiency topics for a given user, as well as filtering their authored posts for those relevant to these topics. Applied to a selected group of users, such as the ones a given user follows on Twitter, this provides a way to focus their time on the content that can be deemed relevant in a professional sense, without having to manually review and consciously discard less proficient opinions.

To this end we evaluated several approaches to model proficiency, including two new models based on implicit user embeddings built on top of User2Vec. While these have been shown to be on-par with existing models in situations where topic-related content can be clearly identified based on query terms, they have shown themselves to be able to capture a wider range of information in situations where the available query terms are sparse. An alternative approach, making use of LDA, has proven itself to be unreliable in order to model proficiency.

\enlargethispage{\baselineskip}
Depending on the task at hand, user embeddings can be used to either model proficiency in the scope of an individual user, by estimating the probabilities of query terms to be authored by a given user, or to model proficiency in a larger scope by relating user-specific probabilities to those obtained on a larger population.

\bibliographystyle{acl_natbib}
\bibliography{paper-lit}

\end{document}